# Intra-day Solar and Power Forecast for Optimization of Intraday Market Participation

Nelson Salazar-Peña[a], Adolfo Palma-Vergara[a], Mateo Montes-Vera[a], María Alejandra Vargas-Torres[a], Rodrigo Hernández Vanegas[a], María Amador[c], Boris Rojas[c], Adriana Salinas[d], Andrés Velasco[d], Alejandra Tabares[b], Andrés González-Mancera[a*]

[a] *Department of Mechanical Engineering, Universidad de los Andes, Bogotá D.C., Colombia*
[b] *Department of Industrial Engineering, Universidad de los Andes, Bogotá D.C., Colombia*
[c] *Strategic Short Term Management, ENEL Colombia, Bogotá D.C., Colombia*
[d] *Renewable Energy Resources Management Col & CA, ENEL Colombia, Bogotá D.C., Colombia*

**Abstract**

The prediction of solar irradiance allows for greater reliability in the energy generation processes of photovoltaic (PV) solar plants and their integration into the power grid. In the Colombian context, the deviation of energy produced compared to the proposed offer by PV solar plants in the intraday market must not surpass an established threshold defined by the national government, or else monetary penalties will take place. Machine learning models are utilized in time series problems and in the prediction of solar irradiance, due to their precision on different time resolutions and prediction horizons. In the present research, Long Short-Term Memory (LSTM), Bidirectional-LSTM (Bi-LSTM), as well as the meteorological data from the PV plant located in El Paso, Cesar, Colombia are used to construct a prediction model for a 6-hour horizon in the future with a 10-minute time resolution. Skill score metrics are employed to choose the more favorable model. The results exposed that the Bi-LSTM possessed significant advantages compared to the other models. Nevertheless, the LSTM model had a similar performance and a shorter training time, proving to be 2/3 faster than the training time of the Bi-LSTM model, which took 18 hours to train. This makes the latter the superior option in terms of computational requirements. Then, the LSTM model was applied to predict solar irradiance, and the predictions within the same hours were averaged to make this an hourly resolution model. Furthermore, the error metrics evaluated consist of the Mean Absolute Error, Root-Mean-Square Error, Normalized Root-Mean-Square Error, and Mean Absolute Percentage Error metrics. The results were compared with the Global Forecast System (GFS) and it was found that the values were similar and both models were capable of representing the behavior of solar irradiance through the day or offered time ranges restricting the behavior of real irradiance in order to predict it with greater precision. Finally, this forecast prediction model is coupled with a power production model that presents an Object-Oriented Programming architecture. With the results obtained by the total power production for a certain horizon, is it possible to accurately issue an energy offer on the intraday market minimizing the penalty costs.

## 1. INTRODUCTION

Fossil fuels are the primary source of energy for electricity generation. However, this source is non-renewable and produces greenhouse gas emissions that contribute to climate change. Annually, humans produce approximately 8 billion metric tons of carbon, with 6.5 billion metric tons from fossil fuel use and 1.5 billion metric tons from deforestation [1]. If this trend continues, the damage will be irreversible, resulting in global temperature increases that will endanger ecosystems and the species that inhabit them. Nevertheless, the need for a reliable network of energy sources is increasingly necessary as demand grows annually worldwide [2]. Given the pressing challenges posed by climate change, the urgency of adopting renewable energy sources has become imperative. To mitigate this problem, renewable energy sources that generate few carbon emissions, such as solar energy, wind energy, geothermal, and hydroelectric plants, are being increasingly utilized. These sources provide an opportunity for sustainable energy production and efficient energy use, as they are replenishable, and their greenhouse gas emissions are significantly lower compared to those of fossil fuels [1].

The solar energy that arrives on Earth in a few minutes is more than 200 times the annual commercial amount used by humanity [1]. Photovoltaic (PV) solar energy is a key technology that plays an important role in achieving clean energy with high accessibility and abundance [3]. However, the fluctuations in solar energy related to weather pose challenges to the stability and reliability of the generated energy and its integration into electrical grids [4]. While the use of storage systems, such as batteries, helps maintain a balanced and stable supply by storing energy during peak production hours and releasing it during low production hours, it is crucial to anticipate this variability to avoid supply issues, ensure reliability, and facilitate management planning. The accurate forecasting of the power generated by PV solar power plants is essential for reliable incorporation into the power generation system.

* Corresponding author.
E-mail: angonzal@uniandes.edu.co

The Colombian short-term energy spot market requires energy generators to submit their energy bids one day in advance (known as the *offer* operation), and through the energy exchange market, where energy offer and demand transactions are executed hourly [5]. The sale in the energy exchange market by PV solar plants is subject to certain tolerances that limit the deviations the plant should have. If a plant promises a certain amount of energy production and exceeds or falls short of this promise beyond the established tolerance values, the plant is monetarily penalized [6]. Predicting changes in irradiance is necessary to achieve greater reliability in processes related to the development and control of solar plants and to avoid monetary penalties due to deviations.

## 1.1. Solar resource forecasting

- *Physical Models*

Variables used for monitoring and predicting solar resources are categorized into four main categories: meteorological, astronomical, geographical, and calendar variables [4]. Meteorological variables are extracted from satellite images, models such as the Global Forecast System (GFS), or meteorological stations; astronomical variables are categorized as theoretical data, which can be calculated through observations of geographical parameters such as longitude, latitude, and altitude. Finally, calendar variables include the day and month of the year and the time of day [4].

Physical models for irradiance prediction are based on Numeric Weather Prediction (NWP) methods. They explore the relationships between irradiance and other meteorological parameters, using atmospheric information such as wind speed, relative humidity, day length, sky images, and satellite images [7], [8]. Additionally, they divide the studied space into a three-dimensional grid and use equations from fluid mechanics, thermodynamics, heat transfer, and parameterizations of other physical processes, defining initial conditions such as the current state of the atmosphere and the spatial domain [9]. NWP models can be classified into two categories: wide-area prediction models, for predicting cloud distribution and solar illumination, and local-area models, used for short-term prediction of PV solar plants [9].

One of the NWP models is the GFS, which is a wide-area prediction model with a base horizontal resolution of 28 km between grid points and a temporal resolution of up to 16 days ahead, making its predictions at four times: 00z, 06z, 12z, and 18z UTC [10]. This is because a climatological reanalysis is performed every 6 hours and the initial state of the GFS is updated [11], [12]. It can increase the spatial resolution to 70 km between grid points.

- *Machine Learning Models*

Machine learning (ML) models have a greater adaptability and flexibility to any kind of problem since they recognize patterns based on empirical data without imposing a model to the data itself. This means that the focus is emphasized on the data instead of being driven by predefined models [13].

In the case of prediction for PV solar plants, by analyzing factors such as solar radiation, temperature, cloud cover, and historical energy generation data, ML models can effectively capture the complex relationships and dynamics underlying solar radiation. This allows us to anticipate fluctuations, optimize energy storage systems, and enhance grid stability, ultimately improving overall efficiency and the integration of solar energy. By leveraging the power of ML algorithms, we can enhance the performance of solar energy systems and facilitate their integration into our existing energy infrastructure.

ML models learn complex patterns and relationships from historical measurements. The computational costs of these models are usually high, as they need to be trained repeatedly to achieve better results. Once the model has been trained, the prediction can be made in less than a second [14]. For problems involving time series, the literature identifies some prospective models to create a prediction model from historical data.

- *Long Short-Term Memory Neural Networks (LSTM)*

LSTM is a type of recurrent neural network designed to effectively capture and model sequential data, such as time series or natural language. Unlike traditional feedforward neural networks, which process data in a single forward pass, LSTM networks have a recurrent structure that allows them to retain and utilize information from previous steps or time points. LSTMs are designed for problems where it is necessary to consider a sequence of information, storing long-term temporal dependencies through memory cells that contain various types of gates. Additionally, they can learn nonlinear relationships [15] [16].

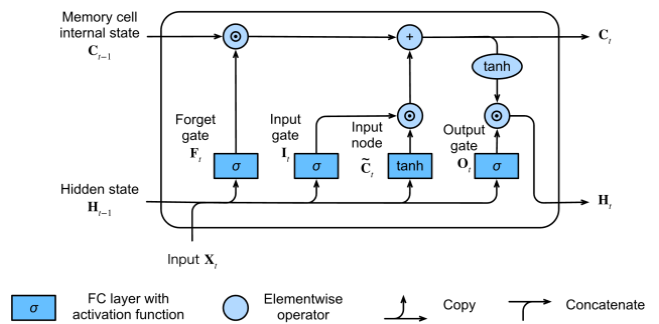

**Figure 1**. LSTM model architecture. Adapted from [15].

- *Bidirectional Long Short-Term Memory Neural Networks (Bi-LSTM)*

Bi-LSTM is the term designated for sequential models that contain two LSTM parallel layers. The first one is used for processing the information in one direction, while the remaining one processes the information in the opposite direction, allowing the model to comprehend and capture the context of the past and future of the input, to realize predictions based on more information [17]. This way, the output of the model depends on the outputs of the forward and backward sequences of the layers.

- *Transformer Neural Networks*

Initially proposed in [18], Transformer models are based on attention mechanisms. That is, unlike recurrent models such as LSTM and Bi-LSTM, which generate a sequence of hidden states based on previous hidden states, attention mechanisms allow modeling dependencies regardless of the distance in the input and output sequence, achieving greater parallelization of processes through encoders and decoders.

Encoders are a grouping of identical layers, each with two sub-layers. The first is part of a multi-head self-attention mechanism, and the second is a simple fully connected layer. At the positional level, these sub-layers are connected through residual and normalization connections. On the other hand, decoders have a similar structure, but they add a third sub-layer, which applies multi-head attention to the output of the encoder. They also add a masked attention layer to ensure that, during the generation of the output sequence, future positions do not influence the attention of the current positions. This masked attention is combined with the fact that the output embeddings are shifted by one position, ensuring that predictions for a specific position only depend on known outputs at previous positions in the output sequence. This is essential for the coherent and accurate generation of the target sequence [8].

## 1.2. Solar PV modelling

Numerous algorithms are available for simulating the performance of PV solar plants. Although these models aim to estimate energy production, they differ in design, assumptions, conceptual approach, mathematical modeling, and data requirements for simulations [19].

In-depth reviews of existing models, detailing their capabilities, contrasts, and limitations, have been conducted [20] [21]. Additionally, Table 1 presents currently active models, along with their intended purposes, years of development, and programming languages.

**Table 1**
Computation tools for modeling PV solar plants. [20] [21].

| Tool | Purpose | Since | Language |
|---|---|---|---|
| SNL PVlib | General modeling of PV systems | 2012 | MATLAB & Python |
| NREL SAM | Modeling of PV systems, solar thermal, and wind farms, including financial analysis | 2004 | C++ |
| PVsyst | Commercial application for modeling PV systems | 1995 | C++ |
| PV*SOL | Commercial application for modeling PV systems | 1998 | C++ |
| Polysun | Commercial application for modeling PV systems | 2009 | C++ |
| SNL Pecos | Performance monitoring of PV systems | 2016 | Python |
| PVMismatch | Estimation of I-V curve mismatches in cells | 2012 | Python |
| CASSYS | General modeling of PV systems | 2015 | Excel & C++ |
| NREL rdtools | Assessment of degradation in PV systems | 2017 | Python |
| pvfactors | Modeling of bifacial irradiance and diffuse shading | 2016 | Python |
| PVFree | API for obtaining PV modeling parameters | 2015 | Python |
| photovoltaic | General modeling of PV systems | 2017 | Python |
| solaR | General modeling of PV systems | 2010 | R |
| feedinlib | Modeling of time series for PV systems | 2015 | Python |

However, the most prominent computational tool for research purposes is PVlib. PVlib is an open-source Python library for modeling PV solar plants [21]. Developed by Sandia National Laboratories (SNL), it continually expands through the Photovoltaic Performance and Modeling Collaboration (PVPMC), a group of PV professionals who share information and work to improve prediction models [22] [23]. Key attributes of PVlib include [23] [24]:

- o Utilizes the Python programming language, offering flexibility and open access for both academic and commercial applications.
- o Supported by a suite of tests and validations (e.g., IEC 61724-1:2023) to ensure library stability and allow model results to be ratified with actual performance data.
- o Enables modeling and analysis of every component in the PV system production chain, with the ability to integrate external Python data analysis libraries.

Additionally, PVlib promotes research and development in the PV industry for several reasons [25]:

- o Facilitates collaboration between researchers and developers, making it easy to adapt new algorithms to existing modeling efforts.
- o Provides validated tools for those who lack the resources to develop them, bridging the gap between industry and research groups and the capabilities of commercial PV analysis software or manual spreadsheets.
- Encourages rapid technological innovation through community contributions, resulting in the adoption of new methodologies, analysis techniques, and best practices.
- Simplifies communication with investors by offering standardized modeling for advanced data analysis, enhancing reproducibility and sharing.

For these reasons, the PVlib Python library is the preferred computational tool for the development of the modeling toolchain for the plant under study: the El Paso PV solar plant of Enel Colombia.

The construction of the computational tool follows the standard PV modeling steps recommended by PVPMC [26] [27]. As can be seen, the modeling of a PV solar plant consists of a sequence of five stages that represent the chain of transformations, energy transport, and losses during the conversion process from the resource to the point of interconnection (POI). The modeling stages are:

1. Technical and electrical design of the PV solar plant.
2. Modeling of the effective irradiance.
3. Modeling of the temperature of the PV cell.
4. Modeling of the DC production.
5. Modeling of the AC production and energy generation.

The first three stages define the input parameters for the electrical modeling of a PV solar plant. These input parameters are: (i) the geographic and technical information of the PV solar plant; (ii) the effective irradiance; and (iii) the PV cell temperature.

- *Technical and electrical design of the PV solar plant*

Regarding the technical and electrical design of the PV solar plant, there are several factors to consider for the proper modeling of the system. These consist of the following parameters:

- Geographical location: Includes the location of the PV solar plant.
- PV panels: Covers the type of technology and conversion efficiency, including the technical information if they are bifacial.
- Inverters: Covers their technical characteristics.
- Design of the PV System: Covers the type of structure to be used: fixed or with a single-axis or dual-axis tracker, and the number of inverters and electrical configuration of PV panels per inverter (i.e., PV panels connected in series and in parallel).
- Global parameters: Includes information on losses due to temperature, losses by the type of structure, losses in the installation, and electrical losses considered up to the POI.

## 2. METHODS

To comprehend how the forecast and power production models are linked, it is necessary to define how the data was analyzed for the former, and which error metrics were utilized to assess the model's performance. Afterwards, the power prediction model is defined and the workflow of how the final computational tool will operate is shown.

### 2.1. Forecast prediction model

A conceptual diagram is exposed in Figure 2. The data used possessed a 10-minute resolution to construct a prediction model consisting of 36 time steps into the future. Additionally, these predictions were grouped in an hourly manner (obtaining 6 time steps in total) with the purpose of acquiring the average of irradiance for the next hour and reducing the prediction to 6 steps with an hourly resolution. The Skill Score (SS), Mean Absolute Error (MAE), Root-Mean-Square Error (RMSE), Mean Bias Error (MBE), and Mean Absolute Percentage Error (MAPE) metrics were evaluated and compared for each model.

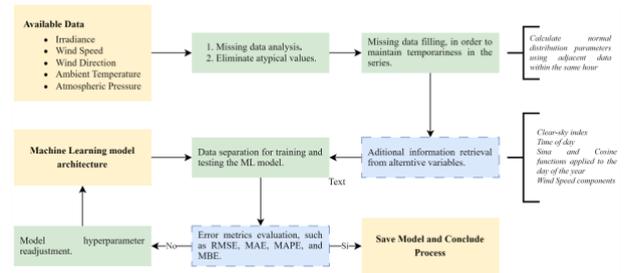

**Figure 2.** ML models construction method based on the data series.

- *Analysis of datasets*

Meteorological data from sensors at the photovoltaic plant location was collected from 2022-02-21 18:00 UTC-5 to 2024-03-03 23:50 UTC-5, with observations recorded every 10 minutes. Data includes GHI (two measurements), ambient Temperature, wind direction, wind speed, and atmospheric pressure. A basic quality control was performed on the data, which allowed to identify of atypical or missing data. Table 2 summarizes the data unavailability for each variable.

**Table 2**
Missing data on the historic series for each variable.

| Variable | Row Quantity | % Empty Rows |
|---|---|---|
| Global Horizontal Irradiance 1 | | 7.47 |
| Global Horizontal Irrdiance 2 | 107112 | 7.77 |
| Ambient Temperature | | 8.52 |

| | |
|---|---|
| *Wind Direction* | 7.13 |
| *Wind Speed* | 7.61 |
| *Atmospheric Pressure* | 7.55 |

The mentioned prediction models work with time series; they learn from these to estimate how the target variable behaves. For this reason, a strategy was sought to fill in the missing data of the available variables to maintain temporality in the training of the models without removing these data from the series. According to [12], [28], [29], the absence of data in time series is a common problem that must be resolved. There is no general rule for imputing data, as there are different filling methods, and some may be better than others depending on the problem, the variable, or the simplicity and ease of implementation. In Machine Learning problems, it is observed that error metrics decrease and the model improves when data are imputed in the time series, regardless of the method used.

To fill in the data, the methodology of [30] was followed, which assumes that data from different days but within the same time slot are normally distributed. Data from the available periods are taken, and the parameters of the normal distribution function are calculated to fill in the missing data. In this way, all data are filled in. Subsequently, outlier values are corrected, such as values below 0 or above a physical limit, for example, ambient temperature above 45°C or below 18°C at the solar plant, when this never occurred during the recorded period. In the case of irradiance, GHI does not exceed 1300 W/m², so values exceeding this limit are also corrected.

- *Training and hyperparameter adjustment*

The data was divided into training, validation, and test sets, with 80%, 10%, and 10% of the samples, respectively. Bayesian optimization was then used to find the optimal hyperparameters for each of the prediction models based on the defined search space and the training data. Finally, the test data was used to evaluate the model's performance, and the t-distribution with 95% confidence was applied to calculate the intervals associated with the prediction error for each horizon.

For the LSTM and Bi-LSTM models, 3 layers of LSTM or Bi-LSTM were used, with the number of internal neurons ranging from 60 to 360 in steps of 30, using a hyperbolic tangent activation function. Additionally, 3 dropout layers were used, exploring dropout rates from 0.2 to 0.5 in steps of 0.1, and the Adam optimizer, where the learning rate was adjusted logarithmically between $1 \times 10^{-6}$ and $1 \times 10^{-2}$. For the Transformer model, the number of encoder and decoder layers was varied between 3 and 6 in step of 1, the number of multi-head attention layers was adjusted between 5 and 20 in step of 3, and dropout rates were also explored.

Each hyperparameter variation was trained for 10 epochs, with 10 variations per model, and each model was run 100 times to identify the best hyperparameter combination for each model.

## 2.2. Power production model

To define the electrical configuration of the El Paso photovoltaic solar plant of Enel Colombia, an architecture was developed with Object Oriented Programming (OOP).

Each class defines a set of elements that make up the photovoltaic solar plant. In addition, they are developed in cascade (see the figure below) to facilitate the representation of, e.g., each array, string box, inverter, and conversion unit (CU). Making a top-down cascade, the OOP architecture is as follows:

```
|-- SYSTEM                          -- class (level 0)
|   |-- name
|   |-- kpc
|   |-- kt
|   |-- kin
|   |-- CONVERSIONUNIT              -- class (level 1)
|       |-- name
|       |-- wire_voltage
|       |-- electrical_resistivity
|       |-- wire_length
|       |-- cross_sectional_area
|       |-- losses
|       |-- SQLTABLE                -- class (level 2)
|       |   |-- database
|       |   |-- table
|       |-- LOCATION                -- class (level 2)
|       |   |-- longitude
|       |   |-- latitude
|       |   |-- altitude
|       |   |-- time_zone
|       |   |-- surface_albedo
|       |-- INVERTER                -- class (level 2)
|           |-- name
|           |-- paco
|           |-- pdco
|           |-- vdco
|           |-- pso
|           |-- c0
|           |-- c1
|           |-- c2
|           |-- c3
|           |-- p_night
|           |-- losses
|           |-- STRINGBOX           -- class (level 3)
|           |   |-- number_of_wires
|           |   |-- electrical_resistivity
|           |   |-- wire_length
|           |   |-- cross_sectional_area
|           |   |-- losses
|           |-- ARRAY               -- class (level 3)
|               |-- modules_per_string
|               |-- strings_per_inverter
|               |-- PANEL           -- class (level 4)
|               |   |-- name
|               |   |-- noct
|               |   |-- technology
|               |   |-- ns
|               |   |-- isc_ref
|               |   |-- voc_ref
|               |   |-- imp_ref
|               |   |-- vmp_ref
|               |   |-- alpha_sc
|               |   |-- beta_oc
|               |   |-- gamma_r
|               |   |-- p_stc
|               |   |-- bifacial
|               |   |-- bifaciality
|               |   |-- mounting
|               |   |-- racking
|               |   |-- degradation
|               |-- TRACKER         -- class (level 4)
|                   |-- with_tracker
|                   |-- surface_tilt
|                   |-- surface_azimuth
|                   |-- axis_azimuth
|                   |-- max_angle
|                   |-- row_height
|                   |-- row_width
```

**Figure 3.** OOP architecture for power generation model.

This OOP architecture allows each component to be represented individually and, therefore, more faithful to real behavior.

The first step to start modeling is to digitally represent the photovoltaic solar plant. This step must be performed when initializing the model and then whenever required.

For this, a plain text file (*architecture_elpaso.json*) is available with which the digital configuration is carried out. For the *System* class there is a method that allows parsing the plain text file (with the structure and syntax present there) to initialize the modeling.

- *Modeling of the effective irradiance*

In accordance with the IEC 61724-1:2021 standard, plane-of-array (POA) irradiance is the sum of direct normal irradiance (DNI) and diffuse horizontal irradiance (DHI) incident on an inclined surface parallel to the PV panels. When measurements of the DNI and DHI components are unavailable, a three-step sequence is required to estimate the POA irradiance. The model that facilitates the estimation of irradiance components is known as the decomposition model, while the model used to estimate POA irradiance from these components is known as the transposition model.

To identify the combination of models that offer greater precision in estimating POA irradiance, proof-of-concept experiments were conducted by combining all available decomposition and transposition models. These combinations were validated by assessing descriptive statistical metrics such as the correlation coefficient ($R^2$) and root mean square error (RMSE) against measured data. The proof of concept led to the recommendation of using the DISC and Perez-Ineichen 1990 models in combination for the estimation and modeling of DNI, DHI, and POA irradiance, yielding values of $R^2 = 0.986$ and RMSE = 3.6%.

To accurately model the effective irradiance, it is necessary to estimate meteorological parameters characteristic of the location. Following the cascade OOP architecture, a *Location* class is associated with each *ConversionUnit* class. Therefore, for each *ConversionUnit* these weather parameters must be estimated. In the case of El Paso photovoltaic solar plant of Enel Colombia, there are twelve (12) instances of the *ConversionUnit* class. The methods used to obtain each parameter are listed below:

- **Solar position:** The implemented method in the Location class uses the algorithm of Reda & Andres [31].

- **Airmass:** The implemented method in the Location class uses the algorithm of Kasten & Young [32].

- **Extraterrestrial DNI:** Many computational tools start with the incident irradiance at the top of Earth's atmosphere, known as extraterrestrial DNI. The implemented method in the Location class uses the algorithm of the National Renewable Energy Laboratory (NREL) because it is more accurate and computationally efficient [33].

- **Clear-Sky index:** The climatic state and its dynamic nature is characterized from the clear-sky index defined by Skartveit & Olseth as the quotient between the POA irradiance and the irradiance in a clear sky condition [34].

- **Relative humidity:** The implemented method estimates the relative humidity from the relationship between water vapor pressure and saturation water pressure.

With the previously estimated parameters, it is possible to estimate the DNI, DHI, and POA using the selected decomposition (DISC) and transposition (Perez-Ineichen) models. However, it is also important to determine the position of the PV solar panel (e.g., when it is coupled to a solar tracker) to then know its orientation with respect to the position of the sun. Additionally, it is necessary to represent the spectral mismatch modifier to depict the effect on the short circuit current of the panel with variations in the spectral irradiance compared to standard conditions. Finally, it is possible to estimate the effective irradiance using the values obtained. The process for conducting each calculation is as follows:

- **Decomposition model:** The DISC decomposition method converts GHI to DNI through empirical relationships between GHI and sky indices. Then, the estimation of the DHI parameter is carried out using the following equation (where $Z$ is the zenith angle) [35] [36].

$$DHI = GHI - DNI * \cos(Z)$$

- **Tracker mounting structure:** Since each *Array* class has an associated *Tracker* class, it is possible to obtain the orientations for each photovoltaic array that composes (i.e., *attribute*) from the *StringBox* class; That is, there is an independent rotation. Additionally, following the cascading sequence, each *StringBox* class is contained in the *Inverter* class and is in turn contained in the *ConversionUnit* class. Because the *ConversionUnit* class has a unique *Location* associated with it, each *Tracker* class of each *ConversionUnit* will also have different orientations with respect to the position of the sun [37].

The plot below shows a single day of operation of the tracker. The y-axis shows the rotation of the tracker from horizontal (i.e. 0 degrees means 'flat' PV panels). In the morning, followers turn at negative angles to face east toward the morning sun. Equivalently, in the afternoon they turn at positive angles to face west towards the sun. At midday the panels are 'flat' because the sun is located in the highest part of the sky.

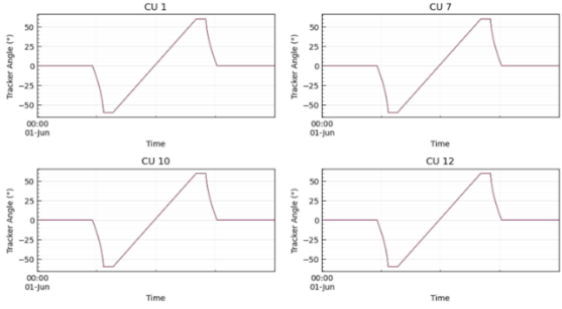

*Figure 4. Daily tracker operation for 4 conversion units.*

- **Spectral mismatch modifier**: The implemented method to obtain the spectral mismatch modifier uses the algorithm of Lee & Pachula [38].

- **Transposition model:** The implemented method uses the algorithm of Pérez-Ineichen [39], which determines the POA irradiance from the direct and diffuse irradiance components relative to the PV panel tilted surface using different parameters such as the angle of incidence, the sky-diffuse component, and the ground-reflected component.

- **Effective irradiance:** The effective irradiance ($G_{effective}$) is estimated by taking into account the incident angle modifier (IAM) and the spectral mismatch modifier (SMM) losses, as follows [40]:

$$G_{effective} = SSM * |(DNI_{POA} * IAM + DHI_{POA})|$$

Where the IAM is calculated from the angle of incidence between the module normal vector and the sun-beam vector.

- *Modeling of the PV cell temperature*

To estimate the temperature of the PV Cell, there are three possible methods. First, the Sandia Array Performance Model represents the PV panel and its mounting and racking through the following equations [41]:

$$T_{panel} = POA * \exp(a + b * WS) + T_{amb}$$

$$T_{cell} = T_{panel} + \frac{POA}{1000 \frac{W}{m^2}} * \Delta T$$

Where $T_{panel}$ represents the back-surface temperature of the models, $a$ and $b$ are empirically-determined coefficients, $WS$ corresponds to the wind speed, $T_{amb}$ is equivalent to the ambient temperature, and $\Delta T$ is the difference in temperature between the cell and the back surface of the module at Standard Testing Conditions irradiance levels [41] [42] [43].

The second method uses the same equations present on the Sandia Array Performance Model but uses the temperature present of the panels to find cell temperatures.

Finally, the third method uses the ambient temperature ($T_{amb}$), the NOCT of the PV panel ($T_{NOCT}$), and the POA irradiance using the subsequent expression:

$$T_{panel} = T_{amb} + \left(\frac{T_{NOCT} - 20\ °C}{800\ \frac{W}{m^2}}\right) * POA$$

When comparing the models, it was observed that the results were close to each other.

- *Modeling of the DC production*

The estimation of the DC production is made from the single-diode equivalent circuit proposed by De Soto et al. [40] and adjusted by Dobos [44], which allows estimating the DC electrical production based on the POA irradiance, the temperature of the PV panel and the technical characteristics of the PV panel.

Since the estimated DC production is for a single PV panel, the following is used to escalate these estimations according to the PV panels arrangement of each *Array* class:

$$P_{DC} = (V_{DC} * PS) * (I_{DC} * PP)$$

Where $PS$ is the number of PV panels connected in series per string in the PV array and $PP$ is the number of strings of PV panels connected in parallel in the PV array.

Like the *Tracker* class, the *Array* class is within each *StringBox* class which, in turn, is contained in the *Inverter* class and then in the *ConversionUnit* class. Because the *ConversionUnit* class has a unique *Location* associated with it, each *Array* class of each *ConversionUnit* will also have different DC production.

The independent DC production for each *Array* contained in the *StringBox* class (in turn contained in the *Inverter* class) is observed in the graphs below.

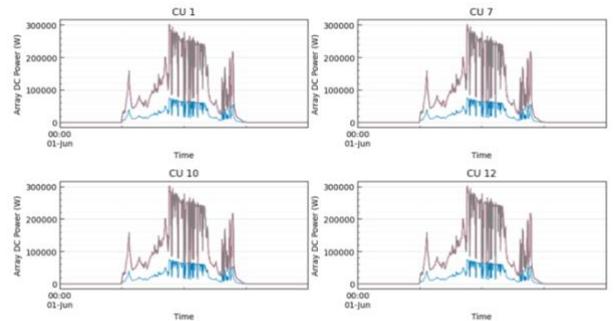

**Figure 5.** DC power production per array for 4 conversion units.

Each Conversion unit is made up of 4 inverters and 8 string boxes are connected to each inverter. Out of the 8 string boxes, 7 string boxes have an array with 30 modules per string and 24 strings per inverter. The remaining string box has 30 modules per string and 6 strings per inverter. Said remaining string box is the blue curve, whose DC production is lower. It is important to mention that except for conversion unit 10, which has 4 inverters with 7 string boxes (all with 30 modules per string and 24 strings per inverter). For the time being, it is assumed that conversion unit 10 has the same configuration mentioned in the previous paragraph and therefore the blue curve with lower DC production is also presented.

- *Modeling of the AC production and energy generation*

To estimate the AC production, the Sandia National Laboratories (SNL) proposes an inverter model which includes the non-linearities of the transformation process and its dependence on temperature, DC production parameters (i.e., voltage and power) and the technical characteristics of the inverter [45].

According to Smets et al. [46] the main quality of the SNL inverter model is that it considers the sources that cause non-linearity between DC and AC power for a given DC voltage. In this way, a dynamic efficiency of the system is achieved, which is more accurate than assuming a linear efficiency. For instance, some of the sources of losses that alter the efficiency of the inverter and that the SNL inverter model takes into account are:

- Self-consumption of the inverter through the parameter $P_{S_0}$.
- Losses proportional to PAC due to fixed voltage drops in semiconductors.
- Ohmic losses by wiring.

The accuracy of the SNL inverter model depends on the data available to determine the performance parameters. Using all required parameters, the model has an approximate error of 0.1% between the modeled and measured inverter efficiency. The results obtained for AC production per inverter are exposed in the figures below.

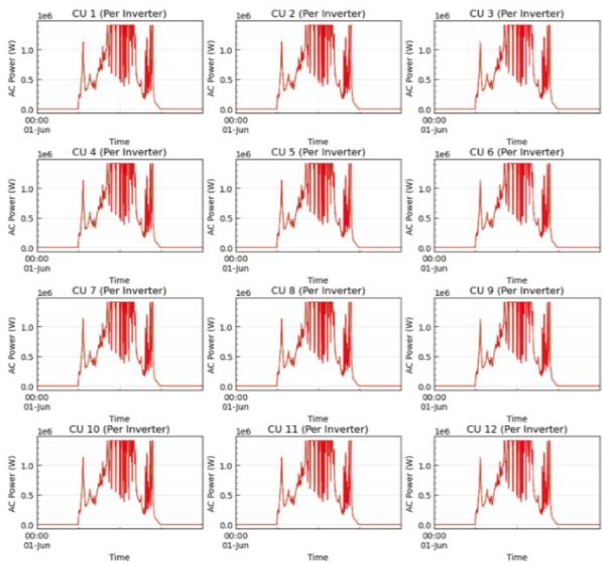

**Figure 6.** AC power production per inverter.

- *Energy generation*

Finally, the energy generated by each Inverter is the sum of the AC power within each time window $t$, multiplied by the factor $r/60$, where $r$ is the resolution of the timestamps in units of minutes.

$$E_{inverter} = \frac{r}{60} * (P_{AC})_t$$

The energy produced per inverter is presented on the subsequent figures.

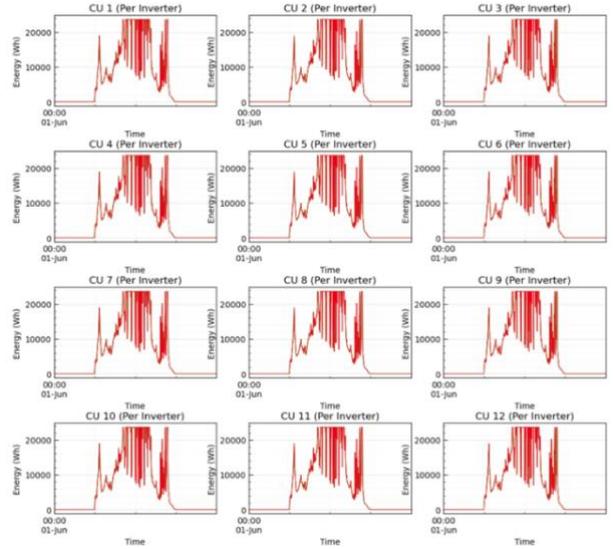

**Figure 7.** Energy production per inverter.

### 2.3. Modelling power production given resource

The final computational tool joins the forecast, power production models, penalties manager, and derived motion wind through the operational flow shown in Figure 8, this has a microservice architecture due to the independence of the services, each of them can be deployed in a component and isolated from others. The above allows fast deployment and segregates the subdomains by their characteristics [47].

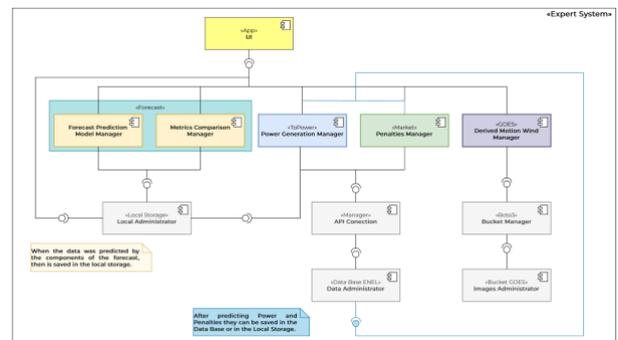

**Figure 8.** Computational tool workflow.

The figure above (See Figure 8) shows the workflow and its components, to estimate the issuing of intraday forecasts, power production, derived motion wind, and penalties given specific operating conditions. These are the services offered by the Expert System. Furthermore, the required information for their calculations can be obtained in two ways: either by reading the information from a database or a file, the second one must be uploaded in the UI. The generated results can be written into a relational database and stored in a warehouse such as Penalties or Power calculations. If required, the data can be downloaded locally (Local administrator).

The structure of the code is presented below, where the required files and directories for proper installation and execution are highlighted in blue (i.e., e*nvironment.yml* and *enel/*). The folder that has * could change in future versions and they are not a service from the source code.

```
1  .
2  |-- build/                      -- dir (despliegue)
3  |-- dist/                       -- dir (compresion)
4  |-- enel/                       -- dir (aplicativo)
5  |   |-- config/                 -- dir (credenciales)
6  |   |-- db/                     -- dir (datos)
7  |   |-- downloads/              -- dir (descargas)
8  |   |-- jobs/                   -- dir (tareas)
9  |   |-- json/                   -- dir (configuracion plantas)
10 |   |-- logs/                   -- dir (registros)
11 |   |-- src/                    -- dir (codigo fuente)
12 |   |   |-- app/                -- dir (servidor web)
13 |   |   |-- forecast/           -- dir (pronostico)
14 |   |   |-- goes/               -- dir (derived motion winds)
15 |   |   |-- manager/            -- dir (gestor datos)
16 |   |   |-- market/             -- dir (penalizaciones)
17 |   |   |-- pi/                 -- dir (servidores externos)
18 |   |   |-- topower/            -- dir (recurso-potencia)
19 |   |-- test/                   -- dir (pruebas)
20 |-- enel_intra.egg-info/        -- dir (requerimientos)
21 |-- img/                        -- dir (imagenes)
22 |-- notebooks/                  -- dir (memorias calculo)
23 |-- .dockerignore               -- file (gestion docker)
24 |-- .gitignore                  -- file (gestion versiones)
25 |-- Dockerfile                  -- file (docker)
26 |-- environment.yml             -- file (ambiente)
27 |-- LICENSE                     -- file (licencia)
28 |-- README.md                   -- file (descripcion)
29 |-- setup.py                    -- file (configuracion)
30 .
```

**Figure 9.** Computational tool structure.

The description of files and directories is shown below:
**Table 3**
Description of files and directories

| Name | Description |
|---|---|
| **db** | Directory where local data is located (e.g, SQLite, Excel, or CSV). For safety concerns, these files aren't shared via repository. |
| **enel** | Directory where the source code (i.e., back-end) of the computational tool is located. It is composed of 4 services, where each one has its own specific subdirectory: |
| *config* | Directory where credentials and sensitive information are located (e.g., username and password for a server or database). For safety concerns, these files are hidden and aren't shared via repository. |
| *app* | Front – end that runs on a local host. |
| *Forecast* | Intraday forecasts prediction model for the primary resource. |
| *manager* | Data manager to get data from a file or the existing API, the last one using request such as GET and POST. |
| *market* | Calculation of the penalties. |
| *pi* | Connection and information download from external sources (e.g., AVEVA PI and Reuniwatt Sky camera). |
| *topower* | Power generation model for photovoltaic solar plants. |
| *goes* | Calculation of the derived motion wind. |
| **jobs** | Directory where the deployment code and specifications for automated execution via cronjob.txt are located. |
| **json** | Directory where the files containing the standardized configuration of the PV plants are stored. |
| **notebooks** | Directory where executable Jupyter notebooks, calculation logs, and validations are located |
| **Dockerfile** | File for build and run the docker of that contains the app. |
| **environment.yml** | File specifying the configurations settings for the application such as: libraries, and their versions required for the correct installation of the environment. |
| **LICENCE** | File displaying the usage license for the computational tool. |
| **README.md** | File where the computational tool is described with the instructions to make the installation. |
| **setup.py** | Equivalent file for environment.yml focused on the PyPI (Python Package Index) installation format. |

### 2.3.1. Source Code (enel) structure and description
**UI Component**
This is the UI developed through the use of the Dash Plotly library. It is continuous deployment will be carried out on a local server. Their purpose is to show the different graphs of the services mentioned before and allow us to analyze them.

**Forecast Prediction Model Manager Component**
Two input parameters are required for the issuing of intraday forecasts for the primary resource. These are:

1. **Data files:** Corresponds to the information of time stamp, GHI $\left(in\ units\ of\ \frac{W}{m^2}\right)$, atmospheric pressure (with $hPa$ as a unit), ambient temperature (in units of °$C$), wind speed $\left(with \frac{m}{s}\ as\ unit\right)$, and wind direction (in units of degrees using the geographical north as 0°) within the last six hours with a 10-minute resolution. The file must present a *.XLSX* extension and the column names must be as represented in Table 4.

**Table 4**
Data file example

| Fecha | GHI | Presion | Temperatura Ambiente | Wind Speed | Wind Direction |
|---|---|---|---|---|---|
| 2022-02-22 00:00:00 | 0.0 | 1003.76 | 24.15 | 1.31 | 276.59 |
| 2022-02-22 00:10:00 | 0.0 | 1003.57 | 24.06 | 1.31 | 226.54 |
| 2022-02-22 00:20:00 | 0.0 | 1003.43 | 24.18 | 1.15 | 284.79 |
| ⋮ | ⋮ | ⋮ | ⋮ | ⋮ | ⋮ |
| 2022-02-22 05:30:00 | 0.36 | 1003.06 | 22.97 | 1.48 | 257.64 |
| 2022-02-22 05:40:00 | 0.38 | 1003.18 | 22.99 | 1.69 | 274.66 |
| 2022-02-22 05:50:00 | 0.40 | 1003.25 | 23.03 | 1.75 | 228.60 |

2. **Model:** Binary file with the trained prediction algorithm parameters. The current extensions for the trained algorithm file are: *.keras (Windows)* and *.h5 (iOS and Linux)*.

This directory contains the implemented LSTM model (see Table 6), the performance metrics, and some calculations required for the predictions. To search for the best hyperparameters the Bayesian optimization was used, the Adam optimizer was selected with a learning rate of *0.00*242, also the loss function was de Mean Absolute Error (MAE).

The output consists of a time series for the next six hours with a 10-minute resolution for the value of GHI.

*pi/* **Folder**

It is possible to connect and download data from the following two external servers:

- **AVEVA PI:** Four input parameters are required:
- **start_date:** Start date in the YYYY-MM-DDThh:mm:ss format (e.g., 2024-05-09T00:00:00).
- **end_date:** End date in the YYYY-MM-DDThh:mm:ss format (e.g., 2024-05-09T23:59:59).
- **freq:** Temporal resolution of the time series to be downloaded (e.g., minute, hourly, daily) following the format accepted by the Pandas library.
- **secondary_database:** Route for the SQLite database where the downloaded data will be located.

The output parameter is the local download of the information measured on-site and stored in AVEVA PI.

- **Reuniwatt Sky camera:** Two input parameters are required:
- **date_start:** Start date in the YYYYMMDD format (e.g., 20240508).
- **date_stop:** End date in the YYMMDD format (e.g., 20240509).

The output parameter is the local download of sky images, and the tabular information measured on-site and stored in the server disposed by Reuniwatt. The downloaded data gathered from both servers will be placed in the downloads/ directory.

**Metrics Comparison Manager Component**

Several models are employed to forecast solar irradiance at the plant, each differing in precision, forecast horizon, and the time of day they are applied. A comparison of these models helps identify the most suitable ones based on the hour of the day, forecast horizon, and season.

Required Input Parameters:
1. **PATH_MODELS:** Path to forecast models' files. Each file must contain the timestamp when the model was used, the forecasted values, and the series ID.
2. **PATH_HORIZONS:** Path to the complementary file for the models. This file should include the series ID, forecast horizon, power plant name, plant sensor information, model names, and resolution IDs.
3. **PATH_IRRADIANCE:** Path to the irradiance data file for the El Paso plant. The file must have timestamps with hourly resolution and corresponding irradiance values.
4. **OPTION:** Metric type to display in the heatmap.
   - 1: Mean Absolute Error (MAE, seen in figure 10)
   - 2: Root Mean Square Error (RMSE, seen in Figure 11)
   - 3: Mean Absolute Percentage Error (MAPE, seen in Figure 12)

The output is a heatmap that displays the selected metric (MAE, RMSE, or MAPE) for the optimal models across different forecast horizons and hours of the day within the reviewed period.

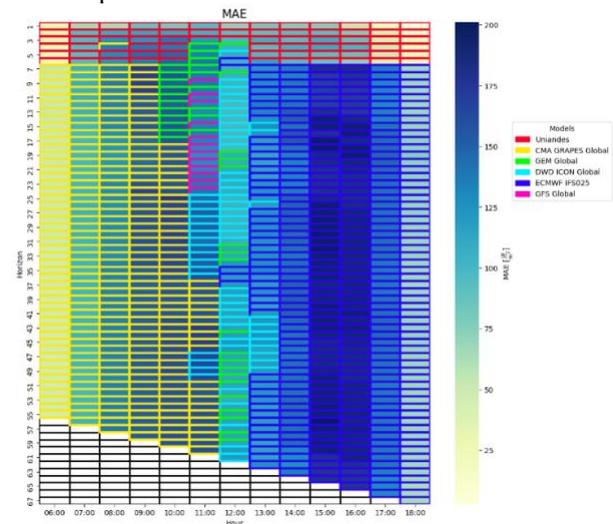

**Figure 10.** MAE for the best models for each forecast horizon and hour of the day

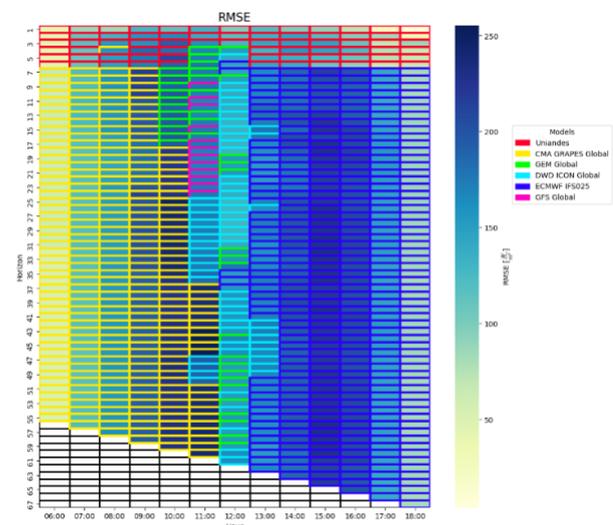

**Figure 11.** RMSE for the best models for each forecast horizon and hour of the day

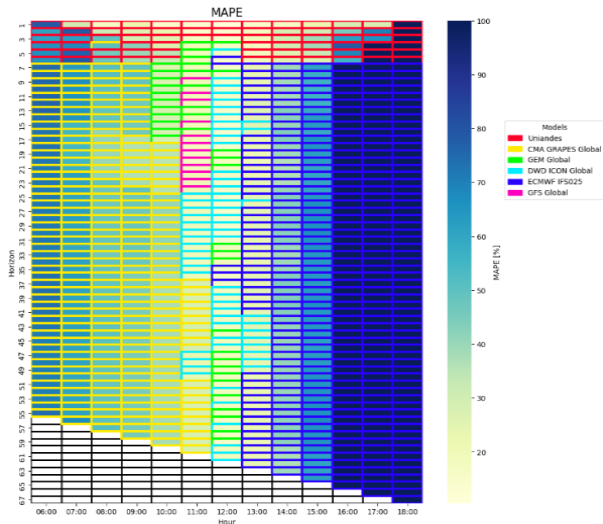

**Figure 12.** MAPE for the best models for each forecast horizon and hour of the day

Figures 10-12 show the comparison between the available forecasting models at the plant for the months of July and August. The outer color of each cell shows the model which performed best in the specified horizon hour, whereas the color inside represents the value of the metric shown.

**Power Generation Component**
To model the power generation of the plant, only one input parameter is required: the plant's electrical configuration. Said configuration is as defined in figure 3 through a plain text file with *extension.json.* It is relevant to state that the required meteorological information will be stored in the *SQL_TABLE* dataset. These time series must include the time stamp, GHI $\left(in\ units\ of\ \frac{W}{m^2}\right)$, and ambient temperature (or panel's temperature) in units of $°C$. Additionally, the following time series are optional: atmospheric pressure (with $hPa$ as unit), relative humidity expressed as a percentage, and precipitation with $cm$ as unit.

The pre-offer, offer, and redispatch operations are made within the present service. For each one of these, the input and output parameters are as follows:

**Pre-offer and offer operations:**
The required input parameters are:

i. **PATH_GFS:** Route for the file of GFS forecast (or *instacast, hourcast,* or *daycast* if the PostgreSQL database is taken). The time series must have the time stamp (with hourly [or greater] temporal resolution), and GHI.
ii. **OPERATION:** Type of operation to perform (i.e., offer or pre-offer).
iii. **DATE_OF_INTEREST:** Corresponding date of interest based on the operation to perform. The date must be expressed in the YYYY-MM-DD format (e.g., 2024-05-09). If it is established as *None,* the date will correspond to the computer's date with one day added for the offer operation, and two days added for the pre-offer operation.
iv. **AVAILABILITY:** Power availability of the plant in units of MW.
v. **PATH_JSON:** Route for the file with the plant's electrical configuration with the *.json* extension.
vi. **PATH_HISTORICAL:** Since the meteorological information obtained from GFS and *Reuniwatt* doesn't contain the time series for ambient temperature, a file with historical information for the photovoltaic plant located in El Paso is disposed (i.e., *el-paso-2020.db* or *el-paso-2020.csv*). With this file, the necessary information is completed. The present parameter defines the route for the file with the historical information.

The output parameter is a file with a *.csv* extension with columns for each period and a row for the day of interest based on the operation. The values correspond to the average AC power of the plant for each hour and are derated in the function of the availability. The results are shown on Table 5 and Figure 13.

**Table 5**
Output parameter for the *topower/* service for pre-offer and offer operations (table).

| Period | 1 | 2 | 3 | … | 11 | 12 | 13 | … | 24 |
|---|---|---|---|---|---|---|---|---|---|
| YYYY-MM-DD | 0 | 0 | 0 | … | 58 | 69 | 69 | … | 0 |

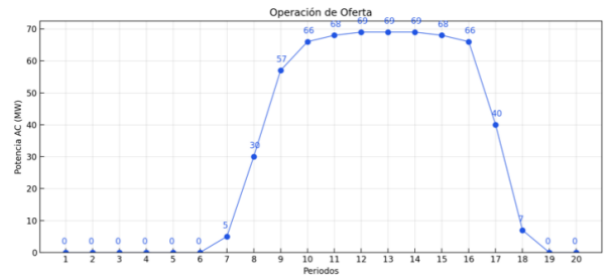

**Figure 13.** Output parameter for the *topower/* service for pre-offer and offer operations (graph).

- **Redispatch operation:**

The necessary input parameters are equivalent to those presented for the pre-offer and offer operations with the exceptions presented below:

i. **PATH_GFS:** Route for the file of GFS forecast previously used for the offer/pre-offer operations. This information represents the power generation for the photovoltaic plant predicted before the day of the operation.
ii. **PATH_DAYCAST:** Route for the *daycast* predictions given by Reuniwatt. Must be *None* if the information originates from the PostgreSQL database.
iii. **PATH_HOURCAST:** Route for the *hourcast* predictions given by Reuniwatt. Must be *None* if the information originates from the PostgreSQL database.

iv. **PATH_INSTACAST:** Route for the *instacast* predictions given by Reuniwatt. Must be *None* if the information originates from the PostgreSQL database.

The output parameter is a file with a *.csv* extension with columns for each period and two rows. The first row corresponds to the GFS information, while the second one represents the predictions established by Reuniwatt. The values are equivalent to the average AC power of the plant for each hour and derated in the function of the availability. The results are shown in Table 6. It is important to note that for the Reuniwatt forecasts, the average power for each period is generated by utilizing all available information from the complete set of forecasts (i.e., *daycast, hourcast,* and *instacast*).

Additionally, the results are also graphed and can be observed in Figure 14. The information derived from GFS is represented with the blue curve along with a ± 5% margin for each period. On the other hand, the red curve exposes the power generation from the information given by Reuniwatt. If this curve falls off the ± 5% margin established for the blue curve, a redispatch operation must be made.

**Table 6**
Output parameter for the *topower/* service for the redispatch operation (table).

| YYYY-MM-DD | 1 | 2 | 3 | … | 11 | 12 | 13 | … | 24 |
|---|---|---|---|---|---|---|---|---|---|
| GFS | 0 | 0 | 0 | … | 68 | 69 | 69 | … | 0 |
| Reuniwatt | 0 | 0 | 0 | … | 69 | 69 | 69 | … | 0 |

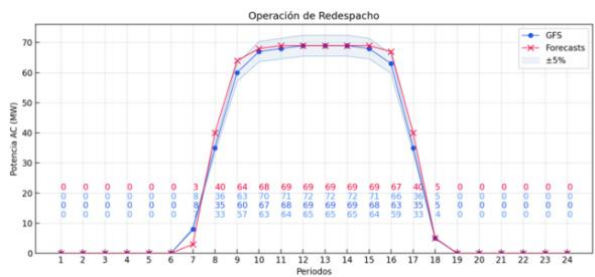

**Figure 14.** Output parameter for the *topower/* service for the redispatch operation.

**Penalties Component**

**Derived Motion Wind Manager Component**

It is of great aid to have a visual representation of what is going on at a satellite level, however, images by themselves do not tell enough information. It is here that the usage of GOES products like Derived Motion Winds (DMW) can become helpful.
This product uses longwave and shortwave infrared measurements to detect features and estimates the movement of each to calculate velocity vectors. To use this product the user must input a date in the format YYYYMMDDhhmm. It is important to note that this product is only available hourly, so the strings corresponding to hh must be numbers from 0 to 24, and the ones corresponding to mm must always be set to 00. By default, this will download and slice the NetCDF files of the DMW and the Cloud Moisture Index products on band 14. A final image displaying the vectors as well as the background clouds will be generated, and the raw files then get automatically deleted.

## 3. RESULTS AND DISCUSSION
### 3.1 Forecasting accuracy for irradiance

Several tests were conducted with different combinations according to the previously defined hyperparameter search space, resulting in the values shown in tables 7, 8, and 9, which best fit the training and validation data.

**Table 7**
LSTM and Bi-LSTM architectures hyperparameters.

| Model | L1 | AF (L1) | D1 | L2 | AF2 (L2) | D2 | L3 | AF(L3) | D3 |
|---|---|---|---|---|---|---|---|---|---|
| LSTM | 240 | tanh | 0.4 | 150 | tanh | 0.2 | 300 | tanh | 0.3 |
| Bi-LSTM | 270 | tanh | 0.4 | 270 | tanh | 0.2 | 360 | tanh | 0.2 |

L: Layer
AF: Activation Function
D: Dropout 1

**Table 8**
Transformer architecture hyperparameters.

| Model | Encoding layers quantity | Decoding layers quantity | Multihead quantity | Neurons in feedforward layer | Dropout |
|---|---|---|---|---|---|
| Transformer | 3 | 3 | 15 | 128 | 0.1 |

**Table 9**
Optimizer architecture used for backpropagation on trained models.

| Optimizer | Alpha, $\alpha$ | Epsilon, $\epsilon$ | Beta 1, $\beta_1$ | Beta 2, $\beta_2$ |
|---|---|---|---|---|
| Adam - LSTM | 0.00242271 | 0.0000001 | 0.900 | 0.999 |
| Adam - BiLSTM | 0.00305146 | 0.0000001 | 0.900 | 0.999 |
| Adam - Transformer | 0.00100000 | 0.0000001 | 0.900 | 0.999 |

The difference in layer size between LSTM and Bi-LSTM seen in table 5 suggests that Bi-LSTM can benefit from a larger number of neurons to capture information in both directions, as it requires more storage and processing capacity. On the other hand, the combination of encoder and decoder layers, along with the number of multi-heads in the Transformer model, makes it a model that can learn from the data and requires less regularization compared to LSTM and Bi-LSTM models, which require higher values.

Furthermore, the Bi-LSTM model has the highest learning rate among the trained models, followed by the LSTM model; however, these are lower compared to values in the literature, such as those used by Michael N et al. [47], which are greater than 0.01. The values found by the Bayesian optimization method allow the model to minimize the loss function in the shortest possible time compared to other learning rates. For the Transformer model, a learning rate of 0.001 was used, as suggested by Vaswani et al. [18], given that the model is more complex due to the number of elements it uses, requiring more time to adjust the internal weights to minimize the error function.

- *10-minute resolution results:*

Test data was used to evaluate each model. The error metrics were calculated excluding nightly hours considering only the hours of operation of the plant (from 6 am to 6 pm).

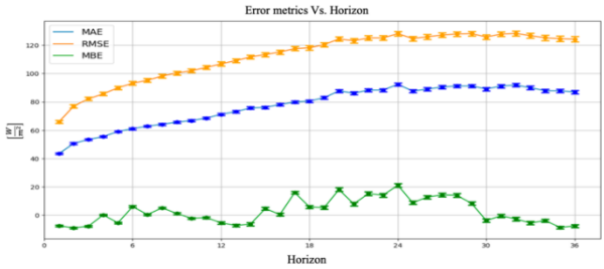

**Figure 15.** Error metrics results for the model with LSTM architecture with 10-minute resolution.

Figure 15 shows that as the prediction horizon increases, errors such as MAE and RMSE increase, while the model bias causes the real irradiance value to be increasingly underestimated. This is consistent with the literature, as it is expected that the model's accuracy with respect to the real irradiance value decreases as the prediction horizon increases.

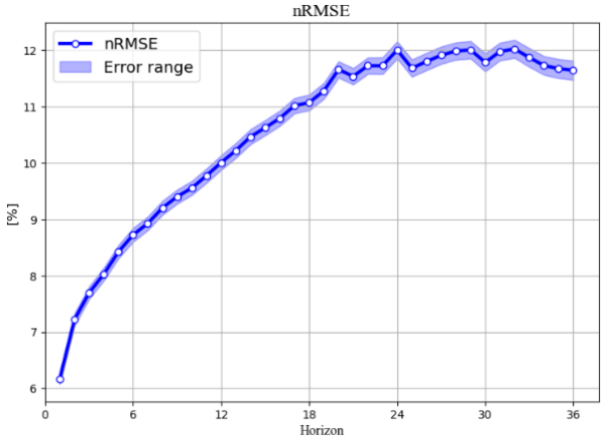

**Figure 16.** nRMSE error metric for LSTM prediction model with 10-minute resolution.

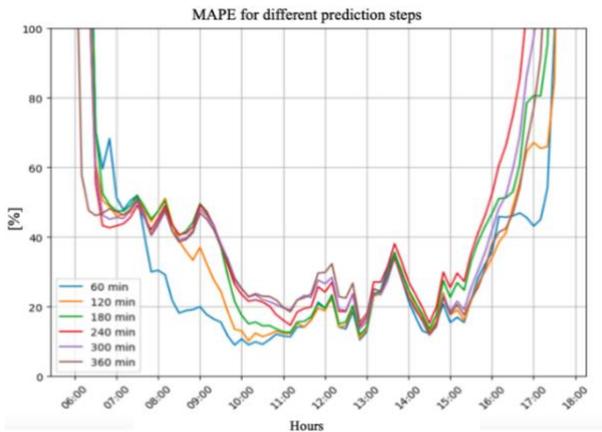

**Figure 17.** MAPE error metric for LSTM prediction model with 10-minute resolution.

In Figure 16, the nRMSE can be observed, which increases along with the prediction horizon, and for the first 12 horizons is less than 10%. On the other hand, in Figure 17, the MAPE metric can be observed for steps from 60 minutes to 360 minutes into the future, showing that for the first step exposed, the percentage deviation is less than or equal to 20% during the hours of highest production in the plants. Additionally, for longer horizons, errors during the hours of highest production can reach up to 40%. The confidence intervals are due to the step at which the results are being evaluated, which is 1, implying that a larger number of predictions are made for the same horizon and time slot.

On the other hand, the Bi-LSTMs, which are a more sophisticated and computationally complex version of LSTMs, performed slightly better than LSTMs in the initial horizons. However, in the later horizons, the LSTMs performed better. Bi-LSTMs require more information to understand the context and temporality of the data series. The hyperparameter optimization time for the LSTM architecture was approximately 6 hours, whereas for Bi-LSTM it was 18 hours. The difference in results is not significant, so until more data is available to better learn the dependencies between variables, it would be better to choose the LSTM model if time has a greater importance compared to other factors.

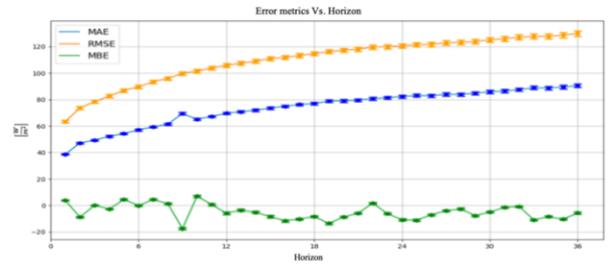

**Figure 18.** Error metrics results for model with Bi-LSTM architecture with 10-minute resolution.

Figure 18 shows that the model bias remains between −18 W/m² and 8 W/m² across the prediction horizons, achieving greater stability compared to the LSTM model, which overestimates the irradiance value by up to 22 W/m². This is due to the Bi-LSTM model's ability to capture the temporality and context of the input and output parameters, resulting in a better generalization of the model.

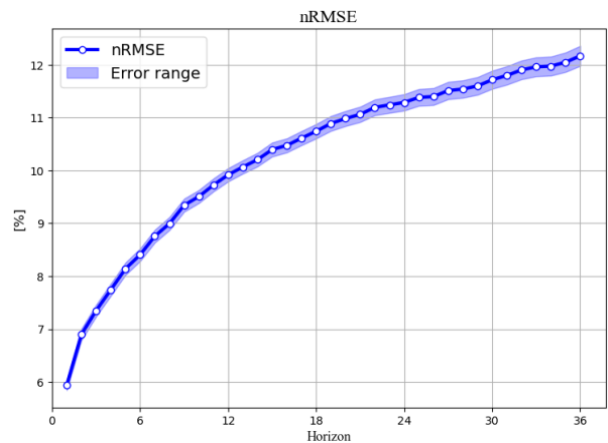

**Figure 19.** nRMSE error metric for Bi-LSTM prediction model with 10-minute resolution.

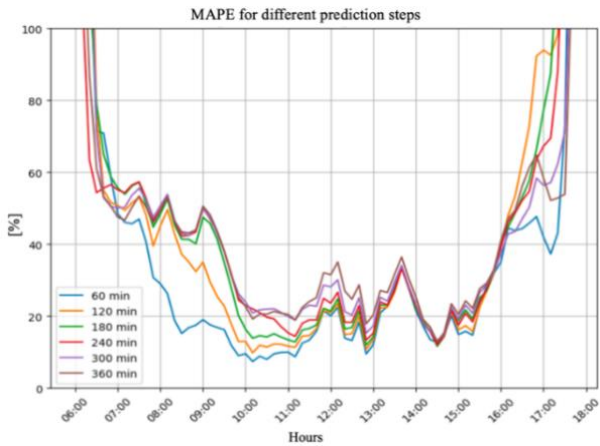

**Figure 20.** MAPE error metric for Bi-LSTM prediction model with 10-minute resolution.

In Figure 19, it is evident that, similar to LSTM, for the first 12 steps it remains below 10%, and it takes longer to reach the value of 12%, which occurs in LSTM starting from the 23rd horizon. However, it does exceed the 12% value in the case of Bi-LSTM. On the other hand, in Figure 20, it is possible to observe that for the 60-minute step during peak production hours, the error remains within 10 to 20%.

Given that the inverters in the plants have a limit on how much they can convert and produce for integration into the grid, the deviations or overestimations of the model at noon indicate that the plant will operate at its nominal power, which is easy to manage in operational terms.

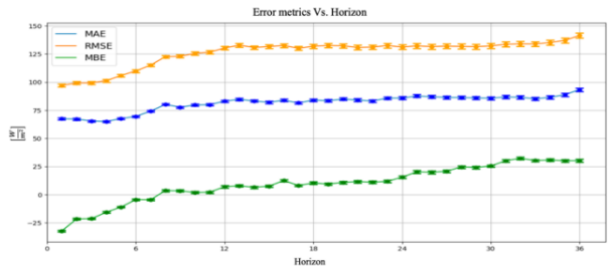

**Figure 21.** Error metrics results for Transformer model with 10-minute resolution.

Figure 21 shows that the Transformer model had the worst performance compared to the LSTM and Bi-LSTM models in nearly all horizons. However, it presents greater stability for each of those horizons, as in the case of MAE, where it remains at the same value without a notable deviation according to the horizon. Nonetheless, it has a greater bias and higher variability.

Since Transformer Neural Networks operate on all the data simultaneously, they do not analyze the problem sequentially, which is why there is no marked increase in error metrics as the horizon increases. However, it is noticeable that they are sensitive to bias, which makes it difficult for the model to generalize and produce consistent results.

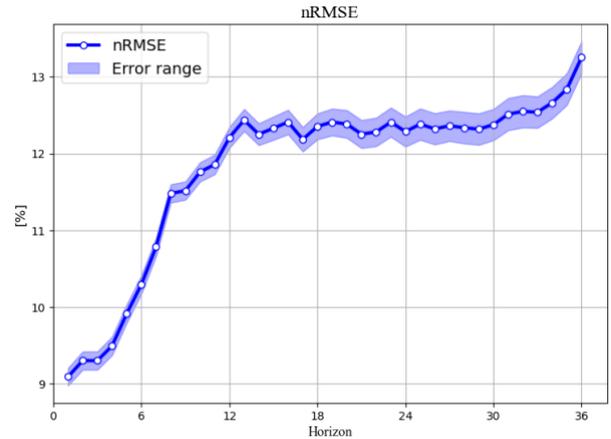

**Figure 22.** nRMSE error metric for Transformer prediction model with 10-minute resolution.

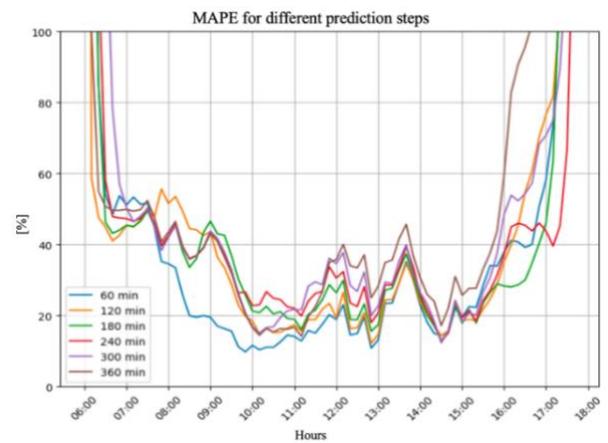

**Figure 23.** MAPE error metric for Transformer prediction model with 10-minute resolution.

As can be seen with the percentage metrics, the error in Figure 22 is greater, and in the first 10 horizons, it shows an abrupt increase compared to the other models where it is a smooth curve. On the other hand, in Figure 23, it is observed that for the 60-minute step, before 8 am, there is an error close to 40%, which did not occur with the other models. Transformer models process the entire data sequence simultaneously, whereas LSTM and Bi-LSTM do so sequentially, which causes the models to identify patterns and relationships in different ways.

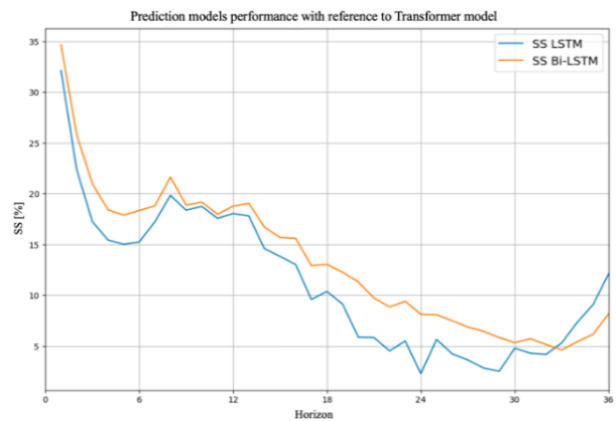

**Figure 24.** Skill Score for LSTM and Bi-LSTM models using the Transformer model as reference.

Figure 24 shows the comparison of the performance between LSTM and Bi-LSTM models regarding the Transformer Neural Network. As previously discussed, the Bi-LSTM model performs slightly better than the LSTM model, except in the last three horizons. Both models outperformed the Transformer, as neither of them had a value below 0% at any point.

- *Hourly resolution results:*

Plants offer their daily production based on hourly production, not in 10-minute intervals. However, training the model allows for better identification of the patterns between the input variables and their respective outputs. The values within the same hour were averaged, thereby approximating the central tendency of irradiance during that period.

The LSTM model, whose hyperparameter tuning time was one-third of that of the Bi-LSTM, was used to predict irradiance at an hourly resolution. Similarly, the values within the same hour were averaged and reported as the irradiance value to be obtained during that hour.

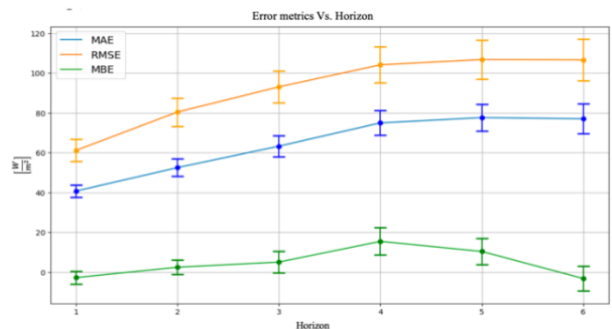

**Figure 25.** Error metrics in hourly resolution after averaging prediction results.

Figure 25 shows the errors by hourly horizon. These were smaller compared to the predictions at 10-minute intervals. For the 3-hour or 180-minute horizon, the MAE metrics differ by 15 W/m², with the hourly prediction being lower and closer to the observed value.

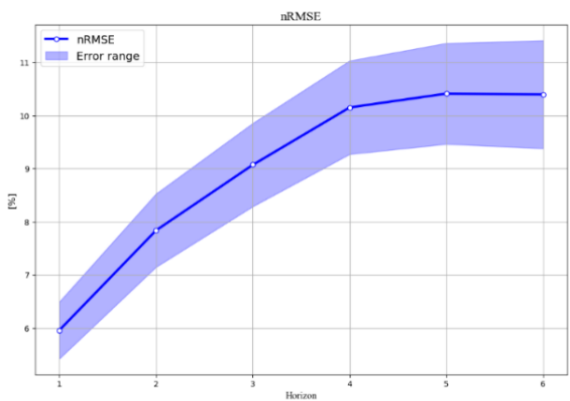

**Figure 26.** nRMSE error metric for prediction model with hourly resolution.

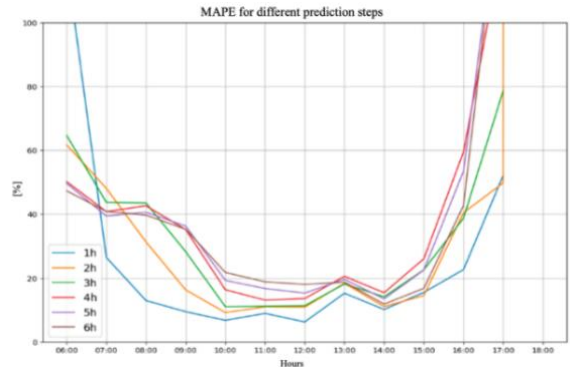

**Figure 27.** MAPE error metric for prediction model with hourly resolution.

In Figure 26, it can be observed that the nRMSE remains below 10% deviations for the first 3 prediction horizons. On the other hand, in Figure 27, we see that the MAPE has a deviation below 20% for all horizons during peak production hours. For the first horizon, it remains between 6% and 20% during most of the hours in which predictions are made. Conversely, for the sixth prediction horizon, it varies between 11.8% and 48% during most hours, excluding 6 am and 6 pm.

Although the error metrics results are mostly above the tolerances permitted by national regulations, the model presents good results considering the error values found in the previously mentioned literature. This allows the model to approach the observed hourly irradiance at various time horizons and provides greater control over the processes within solar plants by having an estimate of the irradiance value that will reach the plant.

The training was conducted with 1 year and 8 months of data. The dependencies between the input parameters and the output will be better learned as more historical data becomes available and the prediction model continues to be trained iteratively

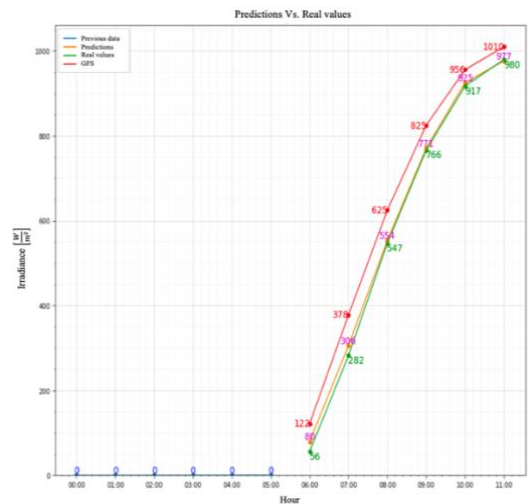

**Figure 28.** Irradiance prediction models results for days with low cloud cover between 06:00 & 11:00.

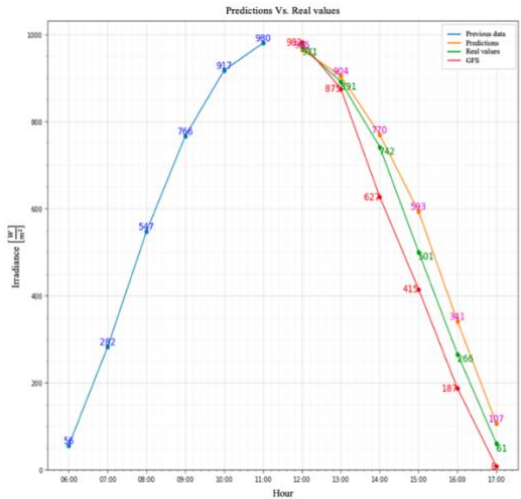

**Figure 29.** Irradiance prediction models results for days with low cloud cover between 12:00 & 17:00.

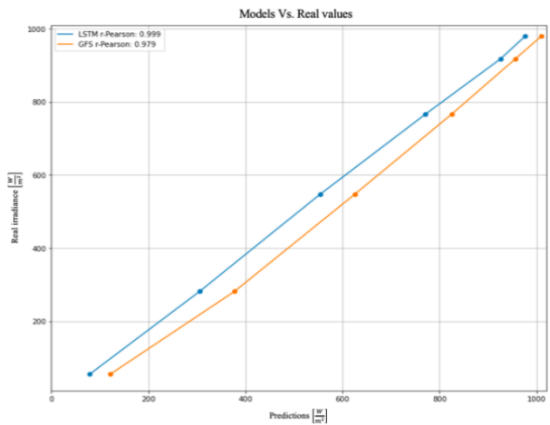

**Figure 30.** Correlation between prediction models and real data for days with low cloud cover between 06:00 & 11:00.

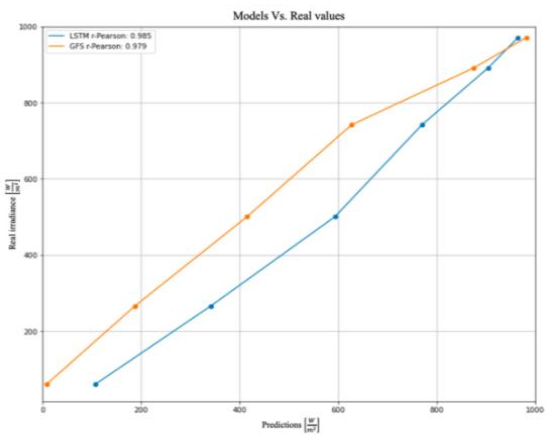

**Figure 31.** Correlations between prediction models and real data for days with low cloud cover between 12:00 & 17:00.

The model performs well on days with low cloud cover, as shown in Figures 28, 29, 30, and 31. The irradiance estimated by the ML model closely approximates the actual values, and it would not have been necessary to run the model multiple times on that day to accurately estimate the average hourly irradiance. The predictions in this case were made using data from 12:00 am to 5:50 am in Figure 28 and from 6:00 am to 11:50 am in Figure 29 on a day with low cloud cover. Additionally, we can see that the LSTM model predictions in Figure 28 were better aligned than those made by the GFS model. The correlation can be observed in Figure 30. A similar pattern occurs in Figure 30 during the initial prediction horizons.

However, in the later horizons, the GFS model provides a better estimate. Even so, the correlation is higher in the ML model, as illustrated in Figure 31. The GFS model uses satellite and sky images for its predictions. In addition to having a database with over 20 years of information, it understands cloud movement and how these phenomena affect irradiance throughout the day, allowing it to estimate irradiance values for the coming hours. Despite this, the ML model has a better correlation coefficient in both cases, making it a competitive alternative given the data used for its training.

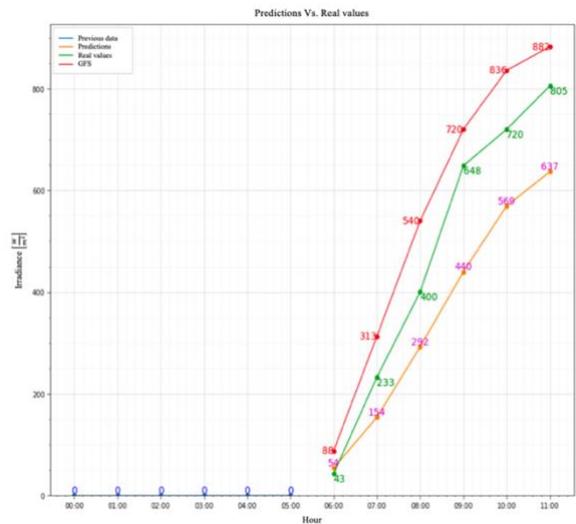

**Figure 32.** Irradiance prediction models results for days with low irradiance between 06:00 & 11:00.

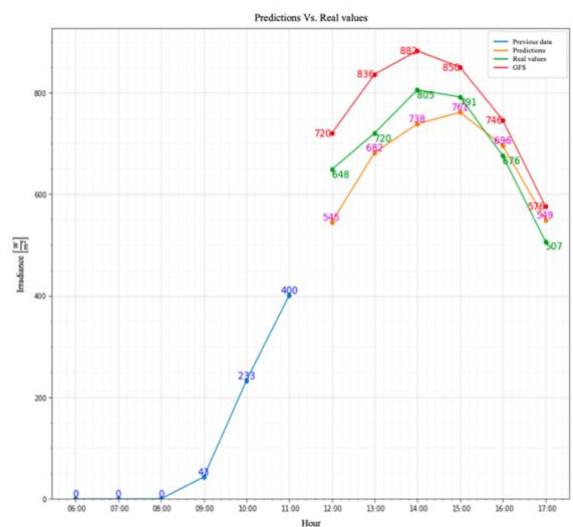

**Figure 33.** Irradiance prediction models results for days with low irradiance between 12:00 & 17:00.

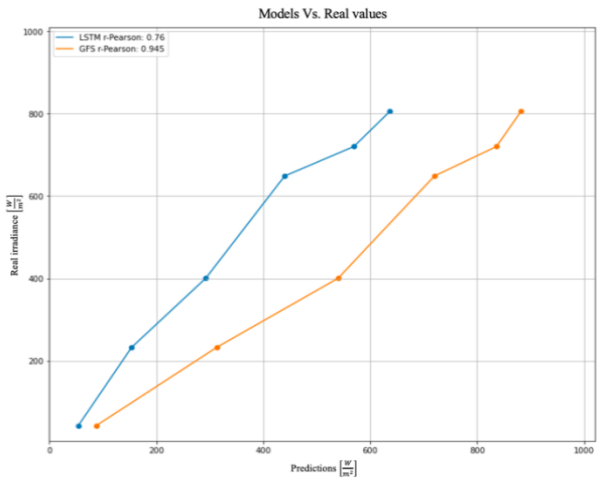

**Figure 34.** Correlation between prediction models and real data for days with low irradiance between 06:00 & 11:00.

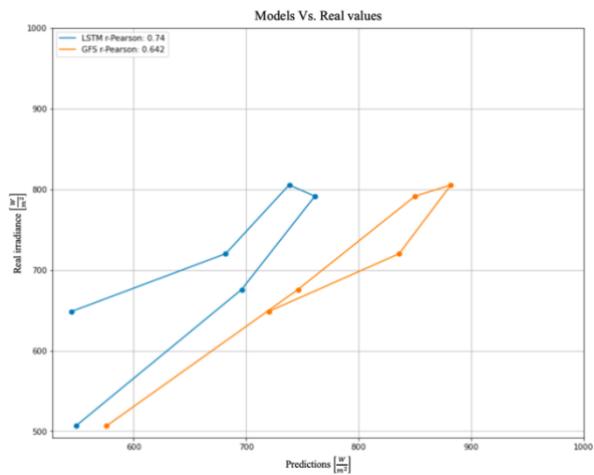

**Figure 35.** Correlation between prediction models and real data for days with low irradiance between 12:00 & 17:00.

In Figure 32, it is possible to identify that during the morning hours, the GFS model better aligns with the observed irradiance values according to the Pearson correlation coefficient shown in Figure 34. However, for the first 3 prediction horizons of the model, the predicted values are closer to the observed values, and it is in the last 3 horizons where the predictions deviate from the observed irradiance values. The MAPE for the first 3 horizons for the LSTM model is 25.58%, 33.91%, and 27.00%, respectively, while for the GFS model, the errors during those hours are 62.96%, 34.33%, and 35.00%. On the other hand, for the last predictions, the MAPE in the LSTM model is 32.10%, 20.97%, and 20.89%, while for the GFS model, it is 11.11%, 16.11%, and 9.56%.

On the other hand, different behaviors are observed during the afternoon hours. Figure 33 shows that the LSTM model aligns closer with the observed data according to the Pearson coefficient presented in figure 35. In this case, the MAPE of the GFS model is lower for the first horizon, while the LSTM's MAPE decreased compared to the first prediction made but remains higher than the GFS at 15.90%. For the remaining horizons, the LSTM model has a MAPE of 5.28%, 8.32%, 3.79%, 2.96%, and 8.28%, respectively, while the GFS model has a MAPE of 16.11%, 9.57%, 7.46%, 10.36%, and 13.60%. It is evident that during the afternoon hours, the LSTM model performs better, even though the GFS model also shows an adequate performance. Both models can model a day with low irradiance and approximating the observed values. However, the LSTM model has the capability to run continuously throughout the day, adjusting to changes in the variables it considers, unlike the GFS model, which requires a 6-hour wait before its next run. Additionally, the training of the models was conducted with data from sensors at the plant, resulting in a higher correlation with its observed behavior. Comparing the predictions of both models allows for the definition of the real irradiance within ranges where it will likely be at each hour, facilitating its prediction.

The Machine Learning model was trained on less than two years of data. Nevertheless, it demonstrates results comparable to those of the GFS. The latter relies on high-performance computing systems to update the states of meteorological variables and is fed by multiple data sources available at different temporal and spatial resolutions. In the case of the Machine Learning algorithm, having access to a larger historical dataset from the plant would allow for better results in both the prediction horizon and the prediction hour. This occurs because the algorithm would be able to use more data to better understand the dependencies between the variables used and irradiance. It is important to emphasize that maintaining data quality and applying regularization techniques is crucial. Without these precautions, the model's performance could deteriorate, or it could overfit the training data, negatively impacting its ability to generalize to new data. Furthermore, the integration of data from other sources, such as additional sensors to measure new variables (relative humidity, air quality, particulate matter), the use of satellite images, or forecasts from other available models could enrich the dataset used for training the model.

If the position of the sensors within the plant is considered, combined models incorporating 2D convolutional layers and time series layers could be integrated. The former would extract spatial information, while the latter would extract temporal information. However, this would increase the complexity and training times of the models.

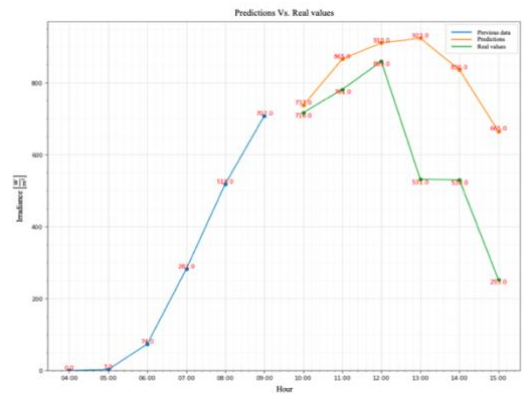

**Figure 36.** Irradiance prediction model results for a day with high variability between 04:00 & 15:00.

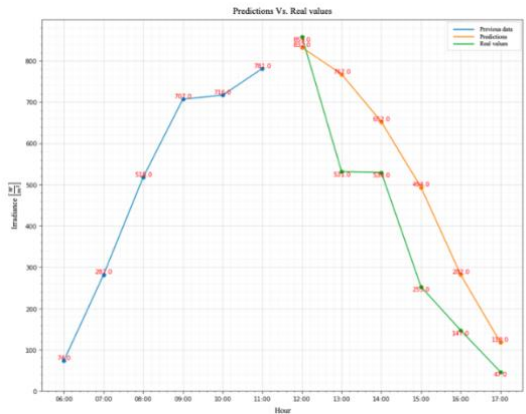

**Figure 37.** Irradiance prediction model results for a day with high variability between 06:00 & 17:00.

In Figure 36, the model closely aligns with the observed value for the first three data points. However, it deviates in the later horizons, indicating that it was not able to anticipate the change reflected by the variables in the previous timestamps. In Figure 37, the prediction decreases for all future periods but diverges from the observed values, except for the first horizon. This suggests that on days with high cloud cover, which can be visually monitored using satellite images or variables like the CSI, it is preferable to run the model every hour to maintain better control over irradiance and its fluctuations throughout the day.

To enhance the model, it should be combined with additional variables, such as satellite images, which can improve its accuracy by providing more detailed information about the atmosphere, cloud behavior, and movement throughout the day. By integrating these images with the model, it is expected to gain the ability to predict cloud behavior over the plant and determine if the plant will be affected by clouds or rain, which would lead to a decrease in irradiance [48].

## 4. CONCLUSIONS

In this project, various short-term solar irradiance prediction models were evaluated using Machine Learning techniques, specifically LSTM, Bi-LSTM, and Transformers, applied to meteorological data from the El Paso solar plant in Cesar, Colombia for the *forecast/* service. The primary objective was to improve the accuracy of predictions for energy supply in the intraday market and reduce monetary penalties due to deviations in energy supply. These models were selected for their ability to handle sequential data and capture complex temporal relationships. The models were trained with meteorological data from the solar plant, such as ambient temperature, global horizontal irradiance, atmospheric pressure, wind speed, and wind direction, along with auxiliary variables derived from this information, such as wind speed components, sine and cosine of the day of the year, and time of day. These variables helped to expand the available dataset and incorporate more physically correlated information with irradiance for model training.

Regarding the models developed, the Bi-LSTM model demonstrated superior accuracy according to the Skill Score metric. However, the LSTM model exhibited comparable performance with a training time of 6 hours, significantly lower than the 18 hours required by the Bi-LSTM, making it a more computationally efficient option. The predictions of the LSTM model were compared with those of the Global Forecast System (GFS), revealing that both models produced similar results. Both successfully managed to capture the daily behavior of solar irradiance, accurately delineating the ranges within which the true irradiance values, as measured by sensors, were found. However, the Machine Learning model has the advantage of being able to run more frequently than the GFS model and can continue training and learning from additional data, whether from the historical records of the initially used variables or from other sources that help expand the plant's database and are correlated with irradiance in some way.

On the other hand, the power production model represented in the *topower/* service considers the data produced by the *forecast/* service to produce a prediction of the possible power generation given certain operation conditions. This power production model utilizes diverse meteorological and technical variables regarding the conditions at the plant. With this information, the model manages to predict the total AC power for a certain hour. With the linkage of the ML forecast model and the power production model, an accurate estimation of generated power can be made which can lead to a more precise pre-offer and offer culminating in lower penalization costs.

## REFERENCES

[1] N. Kh. M. A. Alrikabi, "Renewable Energy Types," *Journal of Clean Energy Technologies*, pp. 61–64, 2014, doi: 10.7763/jocet.2014.v2.92.

[2] K. Kaygusuz, "Biomass as a renewable energy source for sustainable fuels," *Energy Sources, Part A: Recovery, Utilization and Environmental Effects*, vol. 31, no. 6, pp. 535–545, 2009, doi: 10.1080/15567030701715989.

[3] A. E. Gürel, Ü. Ağbulut, and Y. Biçen, "Assessment of machine learning, time series, response surface methodology and empirical models in prediction of global solar radiation," *J Clean Prod*, vol. 277, Dec. 2020, doi: 10.1016/j.jclepro.2020.122353.

[4] Y. Zhou, Y. Liu, D. Wang, X. Liu, and Y. Wang, "A review on global solar radiation prediction with machine learning models in a comprehensive perspective," May 01, 2021, *Elsevier Ltd*. doi: 10.1016/j.enconman.2021.113960.

[5] Comisión de Regulación de Energía y Gas, "RESOLUCIÓN 24 DE 1995," Jul. 1995. Accessed: Nov. 26, 2023. [Online]. Available: https://gestornormativo.creg.gov.co/Publicac.nsf/1c09d18d2d5ffb5b05256eee00709c02/e866f2ef5b7823380525785a007a611d/$FILE/Cr024-95.pdf

[6] Comisión de Regulación de Energía y Gas, "RESOLUCIÓN 60 DE 2019," Jun. 2019. Accessed: Nov. 26, 2023. [Online]. Available: https://gestornormativo.creg.gov.co/Publicac.nsf/1c09d18d2d5ffb5b05256eee00709c02/ca640edbe4b7b5100525842d0053745d/$FILE/Creg060-2019.pdf

[7] Run:ai, "What Is Hyperparameter Tuning and Top 5 Methods." Accessed: Dec. 02, 2023. [Online]. Available: https://www.run.ai/guides/hyperparameter-tuning

[8] A. M. Assaf, H. Haron, H. N. Abdull Hamed, F. A. Ghaleb, S. N. Qasem, and A. M. Albarrak, "A Review on Neural Network Based Models for Short Term Solar Irradiance Forecasting," Jul. 01, 2023, *Multidisciplinary Digital Publishing Institute (MDPI)*. doi: 10.3390/app13148332.

[9] H. Ye, B. Yang, Y. Han, and N. Chen, "State-Of-The-Art Solar Energy Forecasting Approaches: Critical Potential and Challenges," *Front Energy Res*, vol. 10, Mar. 2022, doi: 10.3389/fenrg.2022.875790.

[10] National Oceanic And Atmospheric Administration and National Weather Service, "Numerical Weather Prediction (Weather Models)." Accessed: May 13, 2024. [Online]. Available: https://www.weather.gov/media/ajk/brochures/NumericalWeatherPrediction.pdf

[11] National Oceanic And Atmospheric Administration, "Global Forecast System (GFS) [1 Deg.]." Accessed: May 13, 2024. [Online]. Available: https://www.ncei.noaa.gov/access/metadata/landing-page/bin/iso?id=gov.noaa.ncdc:C00631#:~:text=The%20GFS%20model%20is%20a,its%20performance%20and%20forecast%20accuracy

[12] A. Denhard, S. Bandyopadhyay, A. Habte, and M. Sengupta, "Evaluation of Time-Series Gap-Filling Methods for Solar Irradiance Applications," 2021. [Online]. Available: www.nrel.gov/publications.

[13] P. Wittek, "Machine Learning," *Quantum Machine Learning*, pp. 11–24, Jan. 2014, doi: 10.1016/B978-0-12-800953-6.00002-5.

[14] D. V. Pombo, P. Bacher, C. Ziras, H. W. Bindner, S. V. Spataru, and P. E. Sørensen, "Benchmarking physics-informed machine learning-based short term PV-power forecasting tools," *Energy Reports*, vol. 8, pp. 6512–6520, Nov. 2022, doi: 10.1016/j.egyr.2022.05.006.

[15] O. Calzone, "An Intuitive Explanation of LSTM." Accessed: Mar. 17, 2024. [Online]. Available: https://medium.com/@ottaviocalzone/an-intuitive-explanation-of-lstm-a035eb6ab42c

[16] C. Olah, "Understanding LSTM Networks." Accessed: Mar. 17, 2024. [Online]. Available: https://colah.github.io/posts/2015-08-Understanding-LSTMs/

[17] I. K. Ihianle, A. O. Nwajana, S. H. Ebenuwa, R. I. Otuka, K. Owa, and M. O. Orisatoki, "A deep learning approach for human activities recognition from multimodal sensing devices," *IEEE Access*, vol. 8, pp. 179028–179038, 2020, doi: 10.1109/ACCESS.2020.3027979.

[18] A. Vaswani et al., "Attention Is All You Need," Jun. 2017, [Online]. Available: http://arxiv.org/abs/1706.03762

[19] J. S. Stein, C. P. Cameron, B. Bourne, A. Kimber, J. Posbic, and T. Jester, "A standardized approach to PV system performance model validation," in *2010 35th IEEE Photovoltaic Specialists Conference*, IEEE, Jun. 2010, pp. 001079–001084. doi: 10.1109/PVSC.2010.5614696.

[20] J. Stein and G. Klise, "Models used to assess the performance of photovoltaic systems.," Albuquerque, NM, and Livermore, CA (United States), Dec. 2009. doi: 10.2172/974415.

[21] W. F. Holmgren, C. W. Hansen, J. S. Stein, and M. A. Mikofski, "Review of Open Source Tools for PV Modeling," in *2018 IEEE 7th World Conference on Photovoltaic Energy Conversion (WCPEC) (A Joint Conference of 45th IEEE PVSC, 28th PVSEC & 34th EU PVSEC)*, IEEE, Jun. 2018, pp. 2557–2560. doi: 10.1109/PVSC.2018.8548231.

[22] J. S. Stein, W. F. Holmgren, J. Forbess, and C. W. Hansen, "PVLIB: Open source photovoltaic performance modeling functions for Matlab and Python," in *2016 IEEE 43rd Photovoltaic Specialists Conference (PVSC)*, IEEE, Jun. 2016, pp. 3425–3430. doi: 10.1109/PVSC.2016.7750303.

[23] W. F. Holmgren, C. W. Hansen, and M. A. Mikofski, "pvlib python: a python package for modeling solar energy systems," *J Open Source Softw*, vol. 3, no. 29, p. 884, Sep. 2018, doi: 10.21105/joss.00884.

[24] J. Stein, S. Ransome, W. Holmgren, and J. Sutterlueti, "PV PERFORMANCE MODELLING WITH PVPMC/PVLIB," 2012.

[25] R. W. Andrews, J. S. Stein, C. Hansen, and D. Riley, "Introduction to the open source PV LIB for python Photovoltaic system modelling package," in *2014 IEEE 40th Photovoltaic Specialist Conference (PVSC)*, IEEE, Jun. 2014, pp. 0170–0174. doi: 10.1109/PVSC.2014.6925501.

[26] T. Gurupira and A. Rix, "PHOTOVOLTAIC SYSTEM MODELLING USING PVLIB-PYTHON," 2016.

[27] J. S. Stein, "The photovoltaic Performance Modeling Collaborative (PVPMC)," in *2012 38th IEEE Photovoltaic Specialists Conference*, IEEE, Jun. 2012, pp. 003048–003052. doi: 10.1109/PVSC.2012.6318225.

[28] S. Shan et al., "A deep-learning based solar irradiance forecast using missing data," *IET Renewable Power Generation*, vol. 16, no. 7, pp. 1462–1473, May 2022, doi: 10.1049/rpg2.12408.

[29] T. Kim, W. Ko, and J. Kim, "Analysis and impact evaluation of missing data imputation in day-ahead PV generation forecasting," *Applied Sciences (Switzerland)*, vol. 9, no. 1, Jan. 2019, doi: 10.3390/app9010204.

[30] Á. Pinilla Sepulveda, A. González, A. Pedraza, C. Ramírez, and J. C. Castaño, "Protocolos correspondientes a la resolución 167 de 2017," 2018. Accessed: Mar. 27, 2024. [Online]. Available: https://cnostatic.s3.amazonaws.com/cno-public/archivosAdjuntos/anexo_1_acuerdo_1754.pdf

[31] I. Reda and A. Andreas, "Solar Position Algorithm for Solar Radiation Applications (Revised)," 2000. [Online]. Available: http://www.osti.gov/bridge

[32] F. Kasten and A. T. Young, "Revised optical air mass tables and approximation formula," *Appl Opt*, vol. 28, no. 22, p. 4735, Nov. 1989, doi: 10.1364/AO.28.004735.

[33] W. Holmgren, "irradiance.py tutorial," https://notebook.community/pvlib/pvlib-python/docs/tutorials/irradiance.

[34] A. Skartveit and J. A. Olseth, "The probability density and autocorrelation of short-term global and beam irradiance," *Solar Energy*, vol. 49, no. 6, pp. 477–487, Dec. 1992, doi: 10.1016/0038-092X(92)90155-4.

[35] E. L. Maxwell, "A Quasi-Physical Model for Converting Hourly Global Horizontal to Direct Normal lnsolation," 1987.

[36] M. Mikofski, S. Ayala-Pelaez, and K. Anderson, "PVSC 48 Python Tutorial," https://github.com/PV-Tutorials/PVSC48-Python-Tutorial.

[37] J. Brownson, "Collector Orientation," https://www.e-education.psu.edu/eme810/node/576.

[38] M. Lee and A. Panchula, "Spectral correction for photovoltaic module performance based on air mass and precipitable water," in *2016 IEEE 43rd Photovoltaic Specialists Conference (PVSC)*, IEEE, Jun. 2016, pp. 1351–1356. doi: 10.1109/PVSC.2016.7749836.

[39] R. Perez, P. Ineichen, R. Seals, J. Michalsky, and R. Stewart, "Modeling daylight availability and irradiance components from direct and global irradiance," *Solar Energy*, vol. 44, no. 5, pp. 271–289, 1990, doi: 10.1016/0038-092X(90)90055-H.

[40] W. De Soto, S. A. Klein, and W. A. Beckman, "Improvement and validation of a model for photovoltaic array performance," *Solar Energy*, vol. 80, no. 1, pp. 78–88, Jan. 2006, doi: 10.1016/j.solener.2005.06.010.

[41] D. L. King, W. E. Boyson, and J. A. Kratochvil, "PHOTOVOLTAIC ARRAY PERFORMANCE MODEL."

[42] R. Ross, "Flat-plate photovoltaic array design optimization," in *Photovoltaic Specialists Conference*, San Diego, 1980.

[43] Florida's Premier Energy Research Center, "Cells, Modules, & Arrays," https://www.fsec.ucf.edu/EN/CONSUMER/solar_electricity/basics/cells_modules_arrays.htm.

[44] A. P. Dobos, "An Improved Coefficient Calculator for the California Energy Commission 6 Parameter Photovoltaic Module Model," *J Sol Energy Eng*, vol. 134, no. 2, May 2012, doi: 10.1115/1.4005759.

[45] D. L. King, S. Gonzalez, G. M. Galbraith, and W. E. Boyson, "SANDIA REPORT Performance Model for Grid-Connected Photovoltaic Inverters," 2007. [Online]. Available: http://www.ntis.gov/help/ordermethods.asp?loc=7-4-0#online

[46] A. Smets, K. Jäger, O. Isabella, R. Van Swaaji, and M. Zeman, *Solar Energy: The physics and engineering of photovoltaic conversion, technologies and systems*, 1st ed. Netherlands and Germany, 2016.

[47] C. Richardson, *Microservices Patterns: With Examples in Java*, Manning Publications, 2019. [Online]. Available: https://ieeexplore.ieee.org/book/10280260. [Accessed: 17-Oct-2024].

[48] N. Elizabeth Michael, S. Hasan, A. Al-Durra, and M. Mishra, "Short-term solar irradiance forecasting based on a novel Bayesian optimized deep Long Short-Term Memory neural network," *Appl Energy*, vol. 324, Oct. 2022, doi:10.1016/j.apenergy.2022.119727.

[49] Q. Paletta, G. Arbod, and J. Lasenby, "Omnivision forecasting: Combining satellite and sky images for improved deterministic and probabilistic intra-hour solar energy predictions," *Appl Energy*, vol. 336, Apr. 2023, doi: 10.1016/j.apenergy.2023.120818.